\documentclass[journal]{IEEEtran}
\usepackage[dvipsnames, svgnames,table, x11names]{xcolor}
\ifCLASSINFOpdf
\else
   \usepackage[dvips]{graphicx}
\fi
\usepackage{url}

\usepackage{color}
\usepackage{times}
\usepackage{epsfig}
\usepackage{graphicx}
\usepackage{amsmath}
\usepackage{amssymb}
\usepackage{multirow}
\usepackage{mathrsfs}
\usepackage{array}
\usepackage{ifpdf}
\usepackage{algorithmic}
\usepackage{algorithm}
\usepackage{booktabs}
\usepackage{float}

\usepackage{cite}
\usepackage{subcaption}
\usepackage{changes}
\usepackage{diagbox}
\usepackage{multirow}
\usepackage[colorlinks,
            linkcolor=red,
            anchorcolor=blue,
            citecolor=green 
            ]{hyperref}
            
\begin{document}

\title{Diff-Mosaic: Augmenting Realistic Representations in Infrared Small Target Detection via Diffusion Prior}
\author{Yukai Shi\dag, Yupei Lin\dag, Pengxu Wei, Xiaoyu Xian, Tianshui Chen  and Liang Lin, \IEEEmembership{Fellow, IEEE}

}

\maketitle

\begin{abstract}
Recently, researchers have proposed various deep learning methods to accurately detect infrared targets with the characteristics of indistinct shape and texture. Due to the limited variety of infrared datasets, training deep learning models with good generalization poses a challenge.
To augment the infrared dataset, researchers employ data augmentation techniques, which often involve generating new images by combining images from different datasets. However, these methods are lacking in two respects. In terms of realism, the images generated by mixup-based methods lack realism and are difficult to effectively simulate complex real-world scenarios. In terms of diversity, compared with real-world scenes, borrowing knowledge from another dataset inherently has a limited diversity.
Currently, the diffusion model stands out as an innovative generative approach. Large-scale trained diffusion models have a strong generative prior that enables real-world modeling of images to generate diverse and realistic images. In this paper, we propose Diff-Mosaic, a data augmentation method based on the diffusion model. This model effectively alleviates the challenge of diversity and realism of data augmentation methods via diffusion prior.
Specifically, our method consists of two stages. Firstly, we introduce an enhancement network called Pixel-Prior, which generates highly coordinated and realistic Mosaic images by harmonizing pixels. In the second stage, we propose an image enhancement strategy named Diff-Prior. This strategy utilizes diffusion priors to model images in the real-world scene, further enhancing the diversity and realism of the images.
Extensive experiments have demonstrated that our approach significantly improves the performance of the detection network. The code is available at \href{https://github.com/YupeiLin2388/Diff-Mosaic}{https://github.com/YupeiLin2388/Diff-Mosaic}

\end{abstract}


\IEEEpeerreviewmaketitle

\section{Introduction}
\label {sec:introduction}

\begin{figure}[ht!]
\centering
\includegraphics[width=0.75\linewidth]{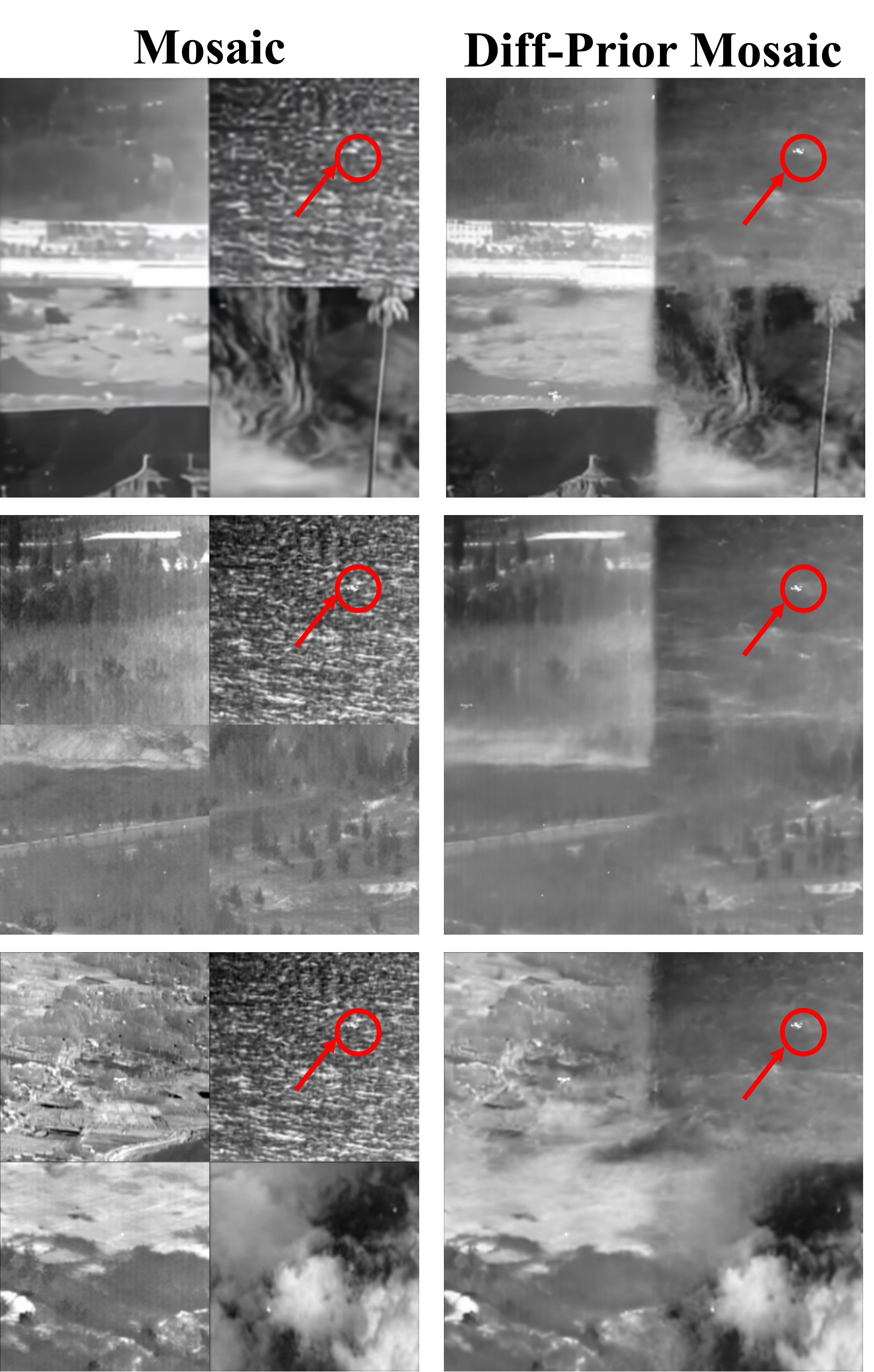}
\caption{
We compared the effects of the traditional Mosaic~\cite{bochkovskiy2020yolov4} with our method. To emphasize the diversity of samples generated by our method, we fix the top-right image of Mosaic when combining rest four images.
The samples generated by the Mosaic exhibit a fragmented quality and fail to resemble a complete image. In contrast, Diff-Mosaic has a uniform distribution and coordinated grayscale. Especially in terms of the infrared small targets marked with red circles, the results of Diff-Mosaic can better enhance the diversity and realism. 
}
\label{fig:Mosaic}
\end{figure}

\IEEEPARstart{S}{ingle}-frame infrared small target(SIRST) detection has a wide range of applications in several fields, such as video surveillance~\cite{xiao2023ediffsr,hu2023cycmunet,xiao2304local}, early warning systems~\cite{deng2016small}, raining scenes~\cite{jiang2024mutual,chen2023towards,jiangfmrnet} and military surveillance~\cite{rawat2020review}. 
Accurately detecting small infrared targets is crucial for ensuring safety, proper navigation, and successful mission execution. However, infrared small target detection faces multiple challenges. Firstly, due to the limited number of pixels occupied by small targets in infrared images, they are easily overwhelmed by complex backgrounds and noise, making them difficult to detect. Secondly, infrared small targets often lack distinct shapes and textures, making their detection more challenging. Additionally, these small targets exhibit weak features in the images, requiring overcoming the impact of lighting variations and other environmental factors for accurate detection. Therefore, achieving accurate detection of infrared small targets in the face of these issues remains a challenging problem.

To achieve infrared small target detection, researchers have proposed many traditional methods. These traditional methods include filtering, local contrast and low-rank-based methods. 
Tophat~\cite{rivest1996detection} and New Tophat~\cite{bai2010analysis} use artificially designed filters to selectively extract visually prominent targets from infrared images. The 2-D least squares filter~\cite{deshpande1999max} predicts the background by estimating the surrounding pixels and detects targets by comparing the difference between the predicted background and the infrared image. 
The Local Contrast Measure (LCM)~\cite{chen2013local}, WSLCM~\cite{han2020infrared} and TCLCM~\cite{han2019local} inspired by the human eye, utilizes local contrast to capture target features. Infrared Patch Image (IPI) models~\cite{gao2013infrared} small target detection as a decomposition problem involving low-rank (background) and sparse signal (target). However, these methods require the selection of appropriate features and carefully tuned hyperparameters, making it difficult to adapt to the diverse noise in real-world scenarios. These methods therefore generate many false alarms. In order to deal with diverse datasets in real-world scenarios, researchers have therefore turned to the use of CNN-based networks.

Unlike traditional methods, CNN-based methods can automatically extract and learn the features of infrared small targets from the dataset, which is more suitable for complex and variable infrared small target detection tasks. Asymmetric contextual module (ACM)~\cite{dai2021asymmetric} accurately detects infrared small targets by integrating shallow and deep feature information. MDvsFA-GAN\cite{wang2019miss} employs an adversarial neural network approach to achieve infrared target recognition by generator learning data distribution. Attentional Local Contrast Network (ALC-Net)~\cite{dai2021attentional} achieves excellent detection performance by embedding low-level detail information into high levels. DNA-Net~\cite{li2022dense} accomplishes accurate target detection using a densely nested interaction module. This module enables progressive interaction between high-level and low-level features, thus maintaining information about small targets in the deep layers. UIU-Net~\cite{wu2022uiu} uses a U-Net in U-Net structure which consists of two U-Nets. The two U-Nets are used for feature extraction and feature fusion respectively, and residual connections and jump connections are used to preserve the details of small targets.

\begin{figure}[t!]
\centering
\includegraphics[width=0.8\linewidth]{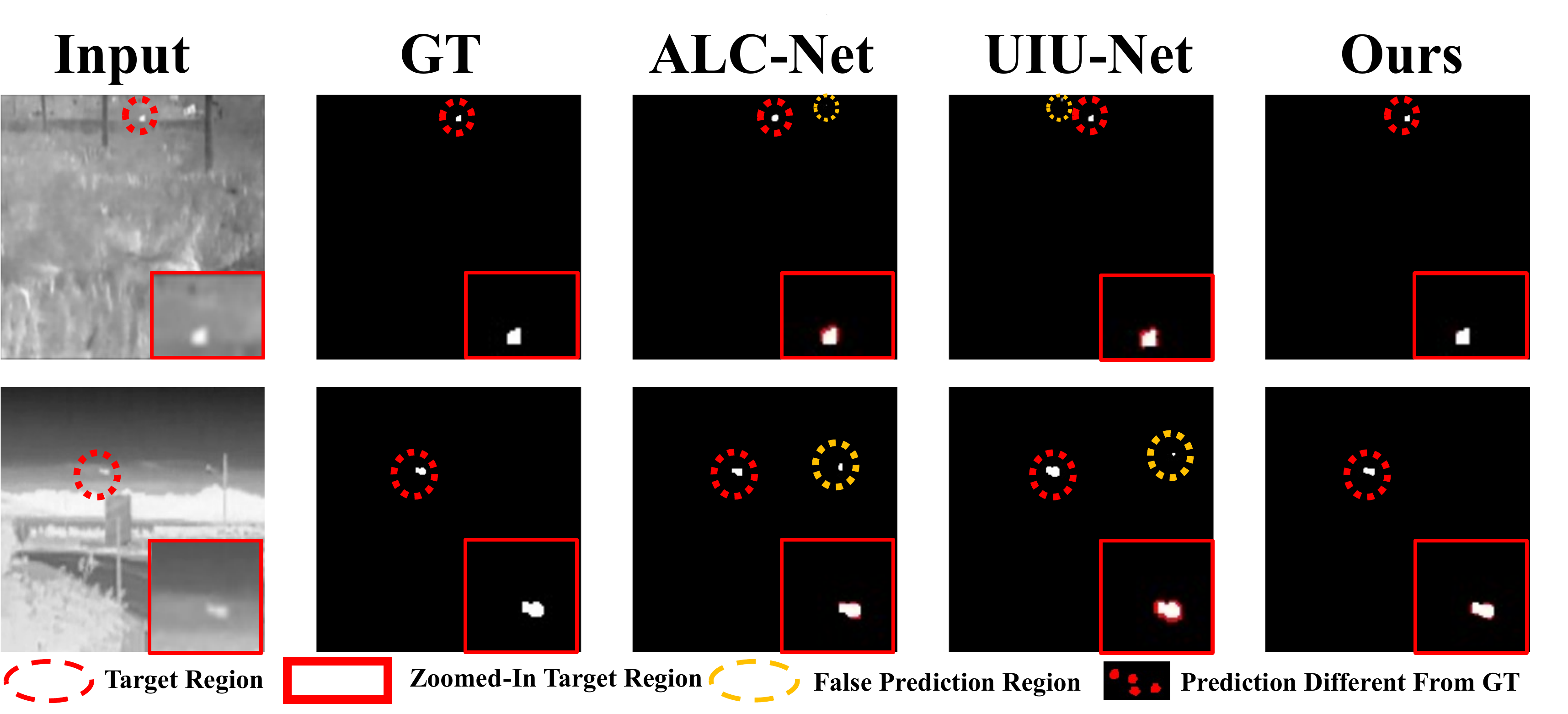}
\caption{We illustrate the visual results of our approach compared to different SIRST detection methods on two SIRST datasets. The target region, Zoomed-In target region, False prediction region, and prediction different from Ground Truth (GT) are annotated with red dashed circles, red boxes, yellow dashed circles, and red pixels, respectively. Unlike other baselines that suffer from false alarms and discrepancies with the ground truth, our approach achieves accurate target detection without false alarms.}
\label{fig:pre_res}
\end{figure}

These methods enhance detection performance by improving the model but do not consider the aspect of proposing a new data augmentation method. 
In SIRST detection, researchers commonly employ typical data augmentation techniques such as Mosaic~\cite{bochkovskiy2020yolov4}, Cut-Mix~\cite{yun2019cutmix}, and Mixup~\cite{zhang2017mixup}. These techniques involve mixing information from different real images to generate new images. However, these data methods are not specifically tailored for infrared datasets, thus posing two challenges in SIRST detection:
\begin{itemize}
  \item \textbf{Diversity.} These mainstream data augmentation methods such as Mosaic, and Cut-Mix generate new images by mixing existing data. However, constrained by the limited number of samples, the diversity of images generated by these data augmentation methods is limited.
  \item \textbf{Realism. }
   As depicted in Fig.~\ref{fig:Mosaic}, pixel distribution of the samples generated by Mosaic is not coherent due to significant contrast variations among infrared images, leading to a lack of realism.
\end{itemize}

The limitations of these data augmentation methods significantly impact the performance of the model.
Improving the diversity and realism of augmented samples is therefore a critical task that requires urgent attention. 
Recently, the Denoising Diffusion Probabilistic Model (DDPM) has attracted considerable interest in low-level domains due to its robust generative prior. The diffusion model uses a powerful generative prior to modeling real-world scenarios based on the distribution of image pixels. The images generated by the diffusion model are significantly improved in detail and realism. Generative Diffusion Prior~\cite{fei2023generative} excels in deblurring and image coloring based on its diffusion generative prior. ControlStyle~\cite{chen2023controlstyle} incorporates textual and visual information via Diffusion Prior, enabling high-quality text-driven stylized image generation.

Inspired by such approaches, in this paper we propose Diff-Mosaic, a novel data augmentation method based on the diffusion model.
Our method employs a diffusion generative prior to integrating real-world information into the images, thereby enhancing the diversity and realism of the generated samples.
Specifically, we propose an enhancement network, Pixel-Prior, based on Mosaic data augmentation, which facilitates the harmonization of image pixels. This network can harmonize images without additional labels, generating high-quality, realistic Mosaic images.
Furthermore, we introduce a diffusion model, Diff-Prior, which integrates real-world information by re-sampling the results of the Pixel-Prior using a diffusion generative prior.
Diff-Prior incorporates real-world knowledge in resampling, which makes the generated samples more realistic and diverse.
Finally, we validated the effectiveness of our method by applying it to the state-of-the-art method in comparison to other detection baselines.
As shown in Fig.~\ref{fig:pre_res}, we compare the visualization results of our method compared to other methods. Our method achieves accurate target detection without false alarms. The contribution of our method is as follows:
\begin{itemize}
  \item In this paper, we introduce an enhanced network that harmonizes images at the pixel level. It generates numerous effective augmented samples without additional labels, overcoming the lack of realism inherent in traditional data augmentation methods.
  \item We introduce an image resampling strategy based on a diffusion model prior. This method uses the diffusion model prior to realistically modeling various aspects of the image, including object shapes, textures, and lighting, thus ensuring the realism and diversity of the generated results. In addition, this approach ensures that our method is the first to introduce a diffusion prior to SIRST.
  \item We apply our proposed data augmentation method on a SOTA baseline. The effectiveness of our method is verified by three evaluation metrics. Ablation studies have shown that each part of the proposed method can improve detection performance.
\end{itemize}

This paper is organized as follows. Section~\ref{sec:related work} briefly reviews related work. In Section~\ref{sec:method}, we describe the flow of the proposed model. In Section~\ref{sec:exp}, we conduct extensive experiments with ablation experiments to demonstrate the reliability of our method. Section~\ref{sec:label} summarizes the paper.

\section{Related work}
\label {sec:related work}

\subsection{Infrared small target detection}
The task of Single-Frame infrared small target (SIRST) detection is to locate abnormal targets in infrared images. Conventional detection methods achieve SIRST detection by comparing the difference between the target and the background. Tophat~\cite{rivest1996detection} and New Tophat~\cite{bai2010analysis} employ manually designed filters to identify high-response in the target region.
The Local Contrast Measure (LCM)~\cite{chen2013local}, WSLCM~\cite{han2020infrared} and TCLCM~\cite{han2019local} detect small targets by assessing the contrast between the central pixel and its surrounding neighborhood at various scales. Infrared Patch Image (IPI) model~\cite{gao2013infrared} decomposes images into low-rank and sparse signals to achieve SIRST detection. However, due to their inability to capture deep features of images, these methods face challenges in detecting anomalous targets within images characterized by complex backgrounds. To address this issue, researchers have employed Convolutional Neural Networks (CNNs) based method to systematically acquire profound features of infrared images through a data-driven approach.

Unlike traditional methods, CNN-based methods do not rely on any prior knowledge about the target and background. Instead, they learn target features directly from the data through the model. 
This advantage makes CNN-based methods widely used in downstream tasks of remote sensing, such as classification tasks~\cite{yu2024hypersinet,zhang2024deep,wu2023three,jha2023mdfs}, segmentation tasks~\cite{miao2023ecae,kou2023lw,gao2023integrating}, and object detection tasks~\cite{shamsolmoali2023efficient}.
CNN-based methods can efficiently process large amounts of data and have strong model-fitting capabilities, thus enabling accurate SIRST detection.
For example, ALC-Net~\cite{dai2021attentional} based on a feature pyramid network proposes an attention local contrast (ALC) network based on the attention mechanism to segment the target region in infrared images. ACM~\cite{dai2021asymmetric} proposes an asymmetric context modulation module and integrates it into U-Net~\cite{ronneberger2015u} and FPN~\cite{lin2017feature}. DNA-Net ~\cite{li2022dense} uses a dense nested interaction module. Multiple interactions of deep semantic features and shallow detailed features are performed through this module to maintain deep infrared small targets. UIU-Net\cite{wu2022uiu} adopts a U-Net in U-Net architecture, consisting of two U-Net components. These two U-Net modules are dedicated to feature extraction and feature fusion, utilizing residual connections and skip connections to retain the details of small targets.

\subsection{Diffusion Prior}
Recently, Diffusion Denoising Probabilistic Models (DDPM)~\cite{ho2020denoising} have made significant progress in computer vision, particularly in image generation. DDPM is divided into two processes: the diffusion process and the reverse process. During the diffusion process, diffusion models progressively introduce Gaussian noise to systematically undermine the input information.
This gradual introduction of noise into the modeling process makes it easier to capture the diversity within images, thereby enhancing the richness and realism of the images generated by DDPM. 
During the reverse process, DDPM then restores the original information by predicting and gradually removing the increased noise. This step-by-step denoising strategy helps the model more effectively reconstruct the original information, particularly demonstrating impressive performance when confronted with complex noise and image degradation scenarios. Thus DDPM has achieved impressive results in image generation.
To further improve the efficiency and quality of image generation, researchers have proposed the Latent Diffusion Model (LDM)~\cite{rombach2022high}.
LDM improves the efficiency of model training and inference by using autoencoder~\cite{kingma2013auto} to compress the information to be learned into a latent space. The large-scale trained LDM model not only generates satisfactory images, but also incorporates real-world information. Therefore, LDM can serve as a powerful generative prior to image enhancement by incorporating real-world information. This diffusion-based generative prior has found applications in various domains. Generative Diffusion Prior~\cite{fei2023generative} and DiffBIR~\cite{lin2023diffbir} leverage the Diffusion Prior to achieve high-quality blind image restoration and image enhancement. ControlStyle~\cite{chen2023controlstyle} integrates textual and visual information using Diffusion Prior, accomplishing high-quality text-driven stylized image generation.


\section{Methodology}
\label {sec:method}

\subsection{Observation}

As discussed in Section ~\ref{sec:introduction}, the scale of publicly available datasets~\cite{dai2021asymmetric,li2022dense} for SIRST detection is limited. Furthermore, the data augmentation methods employed to expand datasets suffer from issues of both limited diversity and a lack of realism in SIRST detection. As shown in Fig.~\ref{fig:Mosaic}, we compare the generation quality of our method with the traditional data augmentation method Mosaic. It can be observed that the images generated by Mosaic look very unnatural at the splices. The four sub-images used for the stitching are not coordinated in terms of brightness and contrast, resulting in an overall lack of realism. In addition, these augmented images are derived from existing infrared datasets, which lack diversity. In order to address this issue, we propose a novel data augmentation network named Diff-Mosaic.

\begin{figure*}[t!]
\centering
\includegraphics[width=0.8\linewidth]{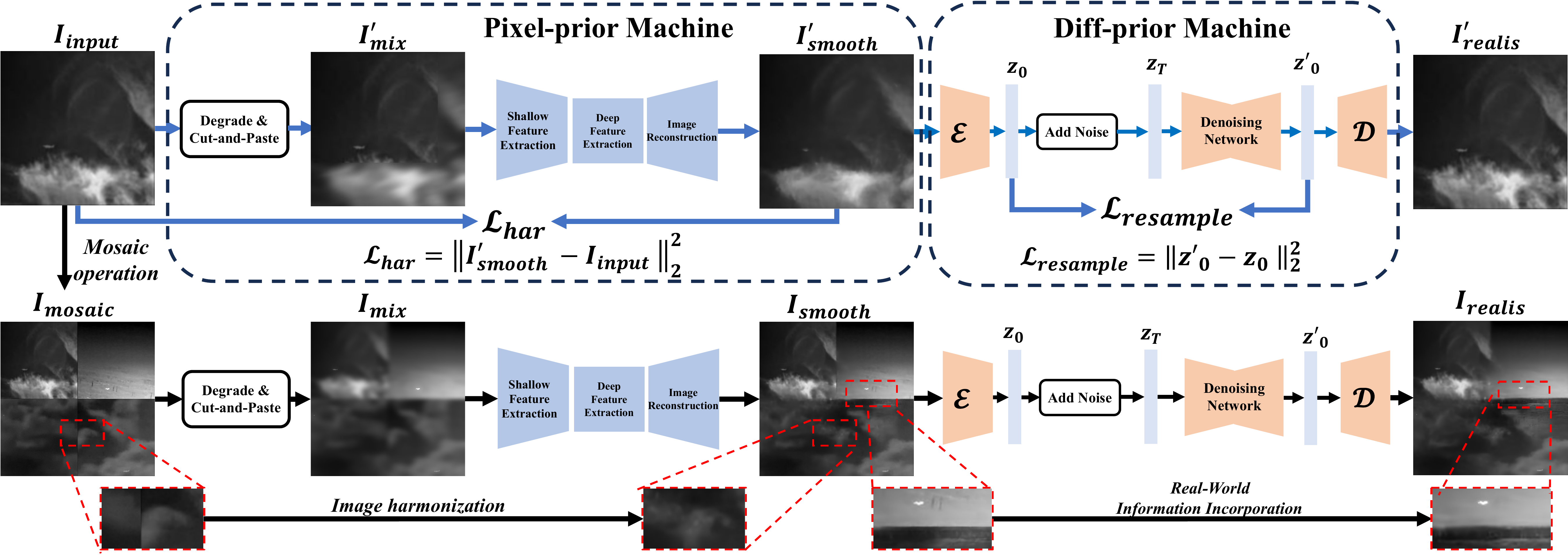}
\caption{Framework overview of Diff-Mosaic. We show the workflow between our training stage and the data augmentation stage of Diff-Mosaic. 
During training, the image $I_{input}$ is subjected to a cut-and-paste operation to get $I'_{mix}$. Subsequently, $I'_{mix}$ is fed into the enhancement network to generate harmonized images $I'_{smooth}$. Finally, $I'_{smooth}$ is fed into the diffusion model for training to get the detail-rich image $I'_{realis}$. 
During the data generation stage, the Mosaic operation is applied to the image $I_{input}$ to yield the Mosaic image $I_{Mosaic}$. Subsequently, $I_{Mosaic}$ is input into the Pixel-Prior machine to generate $I_{smooth}$. Finally, by employing diffusion priors, realistic yet richer representations are integrated into the information of the image $I_{smooth}$ to generate more visually diverse and textured images $I_{realis}$.
}
\label{fig:framework}
\end{figure*}

In this work, we aim to utilize a powerful diffusion generative prior to generating realistic and diverse augmentation samples.
The Diff-Mosaic framework is divided into two aspects: pixel harmonization and detail refinement, the framework is shown in Fig.~\ref{fig:framework}. 
Diff-Mosaic has two stages: the training stage and the data generation stage. In the \textbf{training stage}, the pipeline of Diff-Mosaic is organized as follows:

\begin{itemize}
  \item \textbf{Pixel-prior Machine}: We propose an enhancement network that boosts the quality and realism of images generated by Mosaic.
  First, the input image $I_{input}$ was transformed into $I'_{degrade}$ by applying a degradation module. Then, a cut-and-paste operation was performed to generate a mixed image $I'_{mix}$. Finally, the mixed image $I'_{mix}$ was fed into the enhancement network for training to generate a harmonized image $I'_{smooth}$.
  \item \textbf{Diff-prior Machine}: A powerful generative prior is applied to resample the image $I'_{smooth}$ using the diffusion model, thereby incorporating real-world information. The resampled image $I'_{realis}$ not only contains richer details, but also incorporates real-world knowledge and information from pre-trained diffusion models. Therefore the resampled image $I'_{realis}$ is more realistic and diverse. To make the generated resampled samples more realistic, we fine-tune the diffusion model.
\end{itemize}

In the data generation stage, we use the Mosaic augmentation to generate $I_{Mosaic}$. Subsequently, image degradation and cut-and-paste operations are performed on $I_{Mosaic}$ to generate the mixed image $I_{mix}$. And $I_{smooth}$ is more diverse and has higher image quality than the Mosaic image $I_{Mosaic}$. Finally, we resample $I_{smooth}$ to get the augmented samples $I_{realis}$ via diffusion model. These images will be input into the network as the augmented samples for training. We will describe each step in detail in the following subsections.

\subsection{Mosaic with Pixel-Prior Machine}
\begin{figure}[h]
\centering
\includegraphics[width=0.75\linewidth]{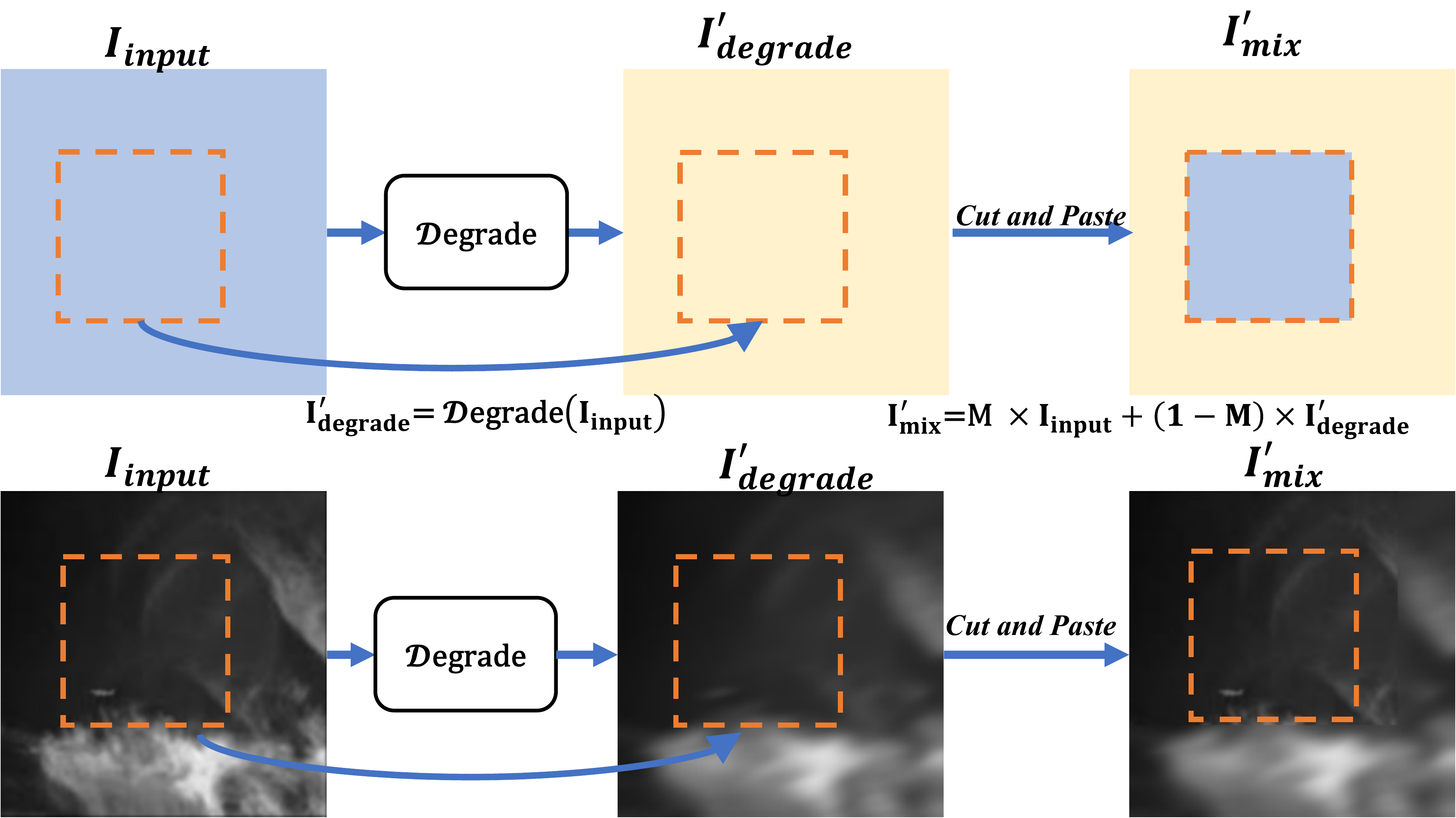}
\caption{Degrade \& Cut-and-Paste process. The image $I_{input}$ is inputted into the degradation module to get $I'_{degrade}$. A random area $M$ is selected from $I_{input}$, cut out, and pasted onto the corresponding area in the degraded image $I'_{degrade}$ to generate the mixed image $I'_{mix}$.
}
\label{fig:degrade}
\end{figure}

In order to solve the problem of uneven distribution existing in Mosaic images and further improve the quality of Mosaic images. We propose Diff-Mosaic to generate image-coordinated Mosaic images. As shown in Fig.~\ref{fig:framework}, in the training stage, we trained a transformer network to enhance the quality of Mosaic images.
Specifically, as shown in Fig.~\ref{fig:degrade}, we conducted degradation and cut-and-paste operations on the input image $I_{input}$ to produce the mixed image $I_{mix}$. This mixed image was then used for training in the enhancement network. The degradation process is expressed as follows:

\begin{equation}
  I'_{degrade} = \mathcal{D}egrade(I_{input})
\end{equation}
 where $\mathcal{D}egrade\left( \cdot \right)$ denotes the degradation module, which employs a variety of degradation sub-modules, including fuzzy, resize, and noise degradation methods. To improve the quality of the reconstructed image, we use the cut-and-paste operation after degradation. Specifically, we cut a part of the region of the original image and paste it into the corresponding region of the degraded image $I'_{degrade}$. The operation is as follows:
\begin{equation}
 \begin{aligned}
  I'_{mix} &= \mathcal{P}(I_{input},I'_{degrade},M_{select}) \\ 
  &= (1-M_{select})\cdot I_{input}+M_{select} \cdot I'_{degrade}
 \end{aligned}
\end{equation}

where $\mathcal{P}(\cdot)$ denotes the cut-and-paste operation, and $M_{select}$ denotes the selection of the region from the original image $I_{input}$.
The Cut-and-paste operation can improve the diversity and difficulties of the data, which makes the recovered images of the model of higher quality.

\begin{figure*}[ht!]
\centering
\includegraphics[width=0.75\linewidth]{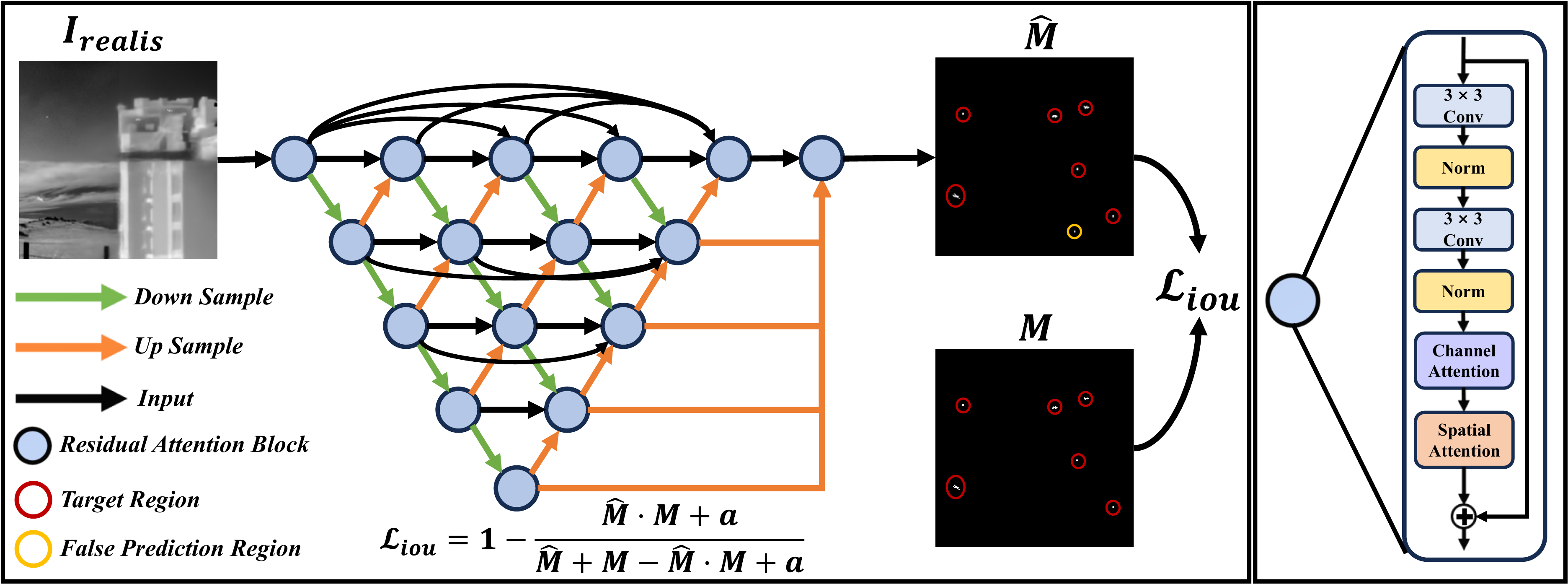}
\caption{Detection framework. We input the image $I_{realis}$ generated by Diff-Mosaic as an augmentation sample into the detection network for training. The backbone network consists of multiple sets of attention modules, each nested attention module consisting of two convolutional layers, a channel attention module, and a spatial attention module. The image input to the densely nested attention module generates features of different scales. Finally, these features are merged to produce the prediction result $\hat{M}$. Red circles denote regions where the target is located, and yellow circles denote regions that were incorrectly predicted. The model closes the gap between $\hat{M}$ and ground truth $M$ by $\mathcal{L}_{iou}$.
}
\label{fig:detection}
\end{figure*}

Then the mix image $I'_{mix}$ is input into the transformer network for reconstruction. 
The transformer network consists of shallow feature extraction, deep feature extraction, and image reconstruction. The input image $I'_{mix}$ is subjected to shallow convolutional layers to extract shallow features. Then it undergoes multiple Residual Swin Transformer Blocks (RSTB) to extract the deep features~\cite{liu2021swin}. Where the RSTB structure consists of multiple Swin Transformer Layers (STL) which collaborate with each other to capture deeper features of the image. Next, we fuse the shallow and deep features to integrate high-frequency and low-frequency information. Finally, the deep features are reduced to the original image space by an upsampling to get the harmonized image $I'_{smooth}$. To optimize the network, we compute the gap $\mathcal{L}_{har}$ between the original image $I_{input}$ and $I'_{smooth}$. The loss $\mathcal{L}_{har}$ is expressed as follows:
\begin{equation}
  \mathcal{L}_{har} = \|I'_{smooth}-I_{input}\|^2_2
\end{equation}

As shown in Fig.~\ref{fig:framework}, compared to Mosaic image $I'_{mosaic}$ , $I'_{smooth}$ effectively solves the distribution inconsistency problem and improves the image quality. 

In the data generation stage, we have adopted the improved Mosaic process. The images are further enhanced to increase the diversity of the augmented samples. Specifically, first, we used Mosaic to stitch image $I_{input}$ with other images in the dataset to form $I_{Mosaic}$. Next, we perform a degradation operation on the image:
\begin{equation}
  I_{mix} = \mathcal{P}\left(I_{Mosaic},\mathcal{D}egrade\left( I_{Mosaic}\right),M_{select}\right)
\end{equation}
Mixed image $I_{mix}$ can improve the diversity of Mosaic, and the mixed image $I_{mix}$ is inputted into the image reconstruction network to generate the Mosaic image $I_{smooth}$ that gets the coordinated Mosaic image.

\subsection{Mosaic with Diffusion-Prior Machine}
Compared to $I_{Mosaic}$, the Pixel-Prior machine-generated image $I_{smooth}$ exhibits a more harmonious and smoother texture. However, there is still a gap between $I_{smooth}$ and the real image. This discrepancy is due to the use of L2 loss in optimizing the enhancement network, which can magnify differences between minor features. Therefore, the Pixel-Prior machine focuses on the uniformity of the overall pixel values while disregarding the details within the image.

To address this problem, we introduce the diffusion model. Unlike traditional methods of data augmentation, the diffusion model employs a powerful generative prior to simulating real-world scenarios based on the distribution of image pixels. This allows diffusion models to incorporate real-world information and generate images with detailed and realistic approximations. In this paper, we use the Latent Diffusion Model(LDM)~\cite{rombach2022high} as a generative prior.

LDM is an advanced diffusion model. It uses pre-trained autoencoders to map images to latent space for learning data distributions. The autoencoder is composed of an encoder $\mathcal{E}$ and a decoder $\mathcal{D}$, where $\mathcal{E}$ encodes input images $I$ into latent code $z$ and $\mathcal{D}$ reconstructs it. In this model, noise is added $T$ times to latent variables $z$, generating Gaussian-distributed codes $z_T$. 
Noise predict network $\epsilon_{\theta}$ is trained to denoise at time step $t$, optimizing with $\mathcal{L}_{ldm}$. $\mathcal{L}_{ldm}$ is calculated as follows:

 \begin{equation}
   \mathcal{L}_{ldm} = \mathbb{E}_{z,t\sim \mathcal{U}(0,1)}\|N_{gaussian}-\epsilon_\theta\left( z_t,t,c\right)\|^2_2
 \end{equation}

where $N_{gaussian}$ is the Gaussian noise and $z_t$ is the latent noise at time step $t$. 
The large-scale trained diffusion model~\cite{rombach2022high} demonstrates proficiency in comprehending various attributes within images, such as object shapes and textures, and is thus able to generate visually appealing images.
To further generate more high-quality and realistic images, we fine-tuned the diffusion model on the SIRST datasets.

As shown in Fig.~\ref{fig:framework}, given $I'_{smooth}$, in the training phase we encode the image $I'_{smooth}$ into the latent space coding $z_0 = \mathcal{E}(I'_{smooth})$ via the encoder $\mathcal{E}$. Then, the $z_0$ model is subjected to a $T$-step denoising into a noise $z_T$ close to a Gaussian distribution, and then $z_0'$ is generated by $T$-step resampling. Finally, the decoder $\mathcal{D}$ generates the resampled image $I'_{realis}$. To bring the results of the diffusion model resampling closer to our dataset, we reduce the gap between $z_0'$ and $z_0$ by $\mathcal{L}_{realis}$. $\mathcal{L}_{realis}$ is expressed as follows:
\begin{equation}
  \mathcal{L}_{realis} = \|z_0-z'_{0}\|^2_2
\end{equation}
During the data generation stage, images $I_{realis}$ resampled by the fine-tuned LDM have much finer details and are very closely aligned with the distribution of the original dataset. In addition, augmentation sample $I_{realis}$ is more diverse and realistic due to the implementation of generative prior.
Notably, this method can capture latent data features without requiring additional labels as input. Therefore, it conveniently produces more realistic and diverse samples for the dataset reconstruction.

In the data generation stage. By employing diffusion prior, we resample $I_{smooth}$ to generate $I_{realis}$. 

As shown in Fig.~\ref{fig:Mosaic}, compared to the Mosaic image, $I_{realis}$ becomes more coherent. In addition, the generated result $I_{realis}$ is very different from other images in the dataset due to the fact that it is extracted information from the real world and generated by the diffusion model, which significantly enhances the diversity of the dataset. The $I_{realis}$ generated in the data generation stage will be involved in the training as the augmented sample.

\subsection{ Infrared small target detection}

To demonstrate the advantages of our proposed data augmentation method, we utilize a standard detection network for inference. The structure of the detection network is shown in Fig.~\ref{fig:detection}. Firstly the image $I_{reails}$ is fed into the residual attention block to extract the features. 
The residual attention block is composed of two convolutional layers, channel attention module, and spatial attention module. These two attention modules are designed to enhance the feature information after passing through two convolutional layers.
In order to fuse different features, the features are continuously upsampled and downsampled into different residual attention blocks to form a feature pyramid. Finally, these features containing different scale information are fused to generate a robust feature map for predicting the result $\hat{M}$. Both the real target result $M$ and the predicted result $\hat{M}$ are binary images. In order to get the results close to ground truth $M$, the network reduces the gap between $\hat{M}$ and the ground truth $M$ by loss $\mathcal{L}_{iou} $, which is expressed as follows:
\begin{equation}
    \mathcal{L}_{iou} =1 -\dfrac{\hat{M}\cdot M+\alpha}{\hat{M}+M - \hat{M}\cdot M +\alpha }
\end{equation}

\section{EXPERIMENTS}
\label {sec:exp}

\begin{figure*}[t!]
\centering
\includegraphics[width=0.75\linewidth]{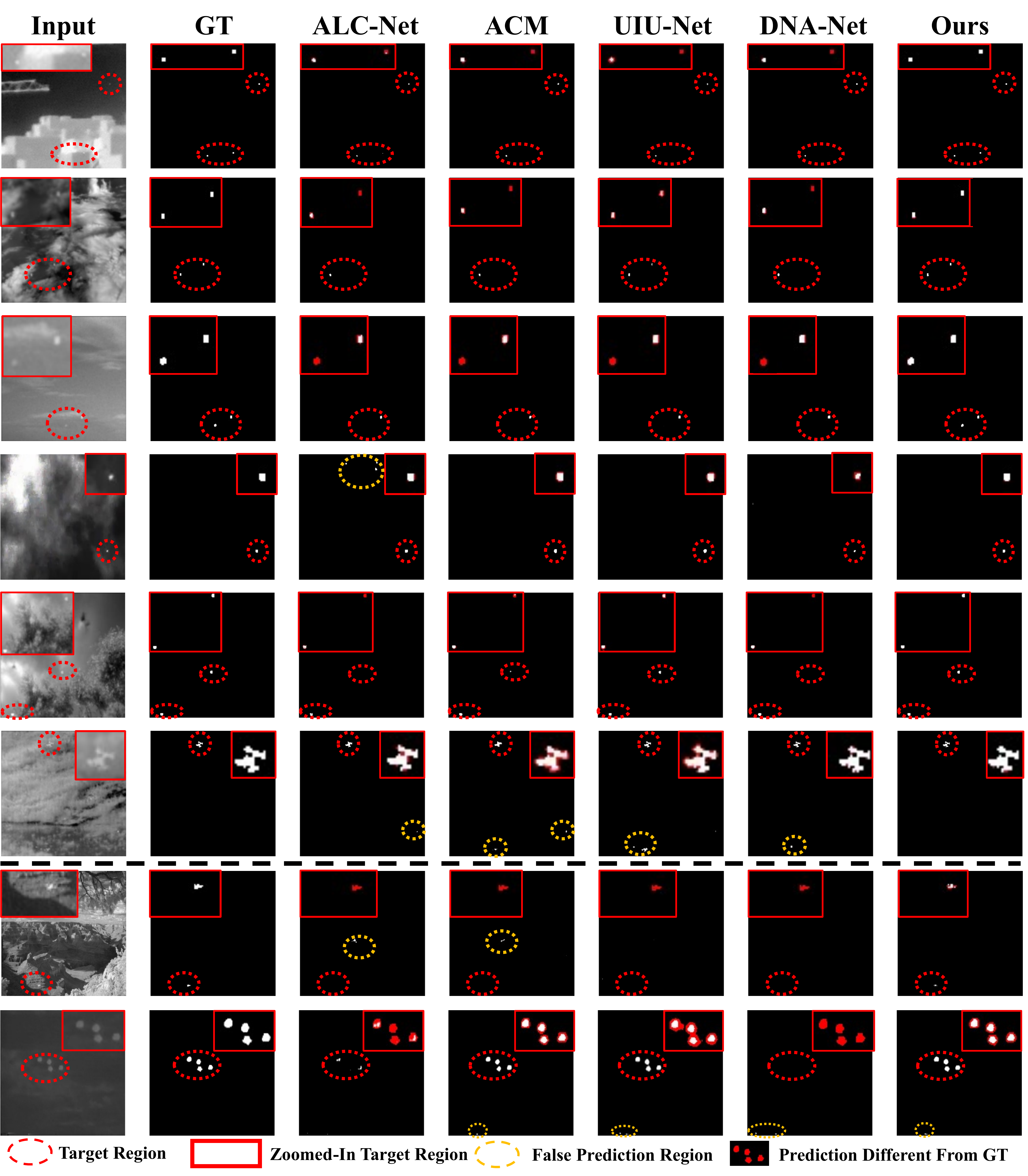}
\caption{
Visualization results. We selected eight images from NUDT-SIRST and SIRST for visualization to compare.
We have circled the area where the small target is located with a red dotted line and zoomed in to better observe the sample. To show the difference between the detection results and Ground Truth (GT), we use red pixels to indicate the gap between the model detection results and ground truth. It can be seen that our results are closer to ground truth.}
\label{fig:visual}
\end{figure*}

\begin{table*}[]
\centering
\footnotesize
\caption{Comparison of IoU, $P_d$ and $F_a$ values between different state-of-the-art methods and our method on SIRST and NUDT-SIRST datasets, where larger values of IoU and $P_d$ indicate higher performance and smaller values of $F_a$ indicate higher performance. We used \textcolor{red}{red} and \textcolor{blue}{blue} to mark the best and second best results and it can be observed that our method achieved the best results.}
\begin{tabular}{cccccccc}
\hline
\multicolumn{2}{c}{\multirow{2}{*}{Method Description}} & \multicolumn{3}{c}{SIRST(Tr=70\%)} & \multicolumn{3}{c}{NUDT-SIRST(Tr=50\%)}    
 \\
 \cmidrule(lr){3-5} \cmidrule(lr){6-8}
\multicolumn{2}{c}{}                  & IoU$(\times10^{-2})$  $\uparrow$ & $P_d$ $(\times10^{-2})$  $\uparrow$ & $F_a$ $(\times10^{-6})$  $\downarrow$ & IoU$(\times10^{-2})$  $\uparrow$ & $P_d$ $(\times10^{-2})$  $\uparrow$ & $F_a$ $(\times10^{-6})$  $\downarrow$ \\ \hline
\multicolumn{1}{c|}{Filtering Based}             & New Tophat~\cite{bai2010analysis}   & 34.05             & 66.91            & 743              & 14.51             & 54.4             & 2086             \\ \hline
\multicolumn{1}{c|}{\multirow{2}{*}{Local Contrast Based}}   & WSLCM~\cite{han2020infrared}     & 20.64             & 60.86            & 1342             & 0.85             & 74.6             & 52391             \\ 
\multicolumn{1}{c|}{}   & TLLCM~\cite{han2019local}    & 9.56             & 56.12            & 3046             & 7.06             & 62.01            & 46.12             \\ \hline \multicolumn{1}{c|}{Low Rank Based}              & IPI~\cite{gao2013infrared}     & 58.2             & 67.1             & 279              & 17.76             & 74.49            & 41.23             \\ \hline
\multicolumn{1}{c|}{\multirow{5}{*}{CNN Based}}        & ALC-Net~\cite{dai2021attentional}   & 70              & 71,71            & 23.91             & 72.13             & 78.81            & 21.22             \\ 
 \multicolumn{1}{c|}{}                     & ACM~\cite{dai2021asymmetric}   & 76.17             & 86.31            & 16.07             & 71.11             & 85.13            & 22.51             \\ 
  \multicolumn{1}{c|}{}                    & UIU-Net~\cite{wu2022uiu}   & 72.03             & \textcolor{blue}{ \textbf{98.1}}             & 26.15             & \textcolor{blue}{ \textbf{89}}              & \textcolor{blue}{ \textbf{98.73}}            & 6.02             \\ 
    \multicolumn{1}{c|}{}                  & DNA-Net~\cite{li2022dense}    &\textcolor{blue}{ \textbf{76.97}}             & 95.41            & \textcolor{blue}{ \textbf{3.54}}             & 88.38             & 97.99            & \textcolor{blue}{ \textbf{4.04}}             \\ \cline{2-8} 
      \multicolumn{1}{c|}{}                & Ours    & \textcolor{red}{\textbf{79.44}}             &\textcolor{red}{ \textbf{99.99}}            &\textcolor{red}{\textbf{ 3.19}}             & \textcolor{red}{\textbf{91.18}}             & \textcolor{red}{\textbf{99.47}}            & \textcolor{red}{\textbf{1.91}}             \\ \hline          
\end{tabular}

\label{tab:exp_res}
\end{table*}

\subsection{Implementation Details}
\subsubsection{Dataset}
To demonstrate the validity of our approach, we chose two datasets for comparison:
\begin{itemize}
  \item Single-frame InfraRed Small Target Detection (SIRST) is an open-source single-frame infrared small target detection dataset that selects images from sequences, which was designated as a public dataset by the University of Arizona in 2020. It contains representative images of 427 different scenes from real-world videos of hundreds of different scenes. These images were taken at shortwave, mid-wavelength, and 950nm wavelengths. They are labeled in five different forms to support the models for the detection task and the segmentation task. The total amount of dataset is 462 and we used 70\% of the SIRST dataset for training and 30\% for testing.
 \item NUDT-SIRST: This dataset is a manually synthesized dataset of five main background scenes including city, field, highlight, sea, and cloud. Each image is synthesized from a real background with different targets (e.g., point, dot, and extended targets) with different SCRs and rich poses. The total amount of data is 1362 and we divide the dataset into 50\% for training and 50\% for testing. 
\end{itemize}

\subsubsection{Implementation Details}
To achieve data augmentation for extracting knowledge from real scenes, we trained the Diff-Mosaic.
For the degradation part in the Pixel-Prior machine, we used second-order degradation. The blur-resize-noise process is repeated twice for the image input~\cite{lin2023diffbir}. For the reconstruction network, we set the batch-size to 4, the training epoch to 100, and the learning rate to 0.001. For the Diff-Prior machine, we fine-tune the diffusion model 100 epoch and use the Adam optimizer, and set the learning rate to $10^{-4}$.
In this experiment, we generated 400 augmented samples on the NUDT-SIRST dataset and 100 augmented samples on the SIRST dataset via Diff-Mosaic. The detection network was trained with batch-size 40 and an epoch number of 3000.
\subsubsection{Comparison Method}
To demonstrate the effectiveness of our method, we compare our method with the SOTA method, including traditional methods (New Tophat,~\cite{bai2010analysis} WSLCM~\cite{han2020infrared}, TLLCM~\cite{han2019local}, IPI~\cite{gao2013infrared}) and CNN-based methods(ACM~\cite{dai2021asymmetric}, ALC-Net~\cite{dai2021attentional}, UIU-Net~\cite{wu2022uiu} and DNA-Net ~\cite{li2022dense}).
For a fair comparison, all CNN-based methods were trained with 3000 epochs, and the other configurations were consistent with the original paper. ACM and ALC-Net were implemented using Mxnet, and DNANet and UIUNet were implemented using PyTorch 1.10 and run on an NVIDIA Geforce RTX 3090.

\subsection{Evaluation Metrics}

To evaluate the detection performance of the CNN-based approach, we use three different evaluation metrics to assess the network performance.
\begin{itemize}
  \item Intersection over Union (IoU): IoU is a classical pixel-level semantic segmentation evaluation metric to describe the algorithm's contour description capability. It is defined as the ratio of intersection and concatenation area between predicted values and labels as follows:
  \begin{equation}
    IoU = \dfrac{A_{inter}}{A_{Union}}
  \end{equation}
  where $A_{inter}$,$A_{Union}$ denotes the interaction region and union region, respectively.
  \item Probability of detection($P_d$ ): The probability of detection  is a target-level evaluation metric. It measures the ratio of the number of correctly predicted targets $T_{correct}$ to the number of all targets $T_{All}$. The definition of $P_d$ is as follows:
  \begin{equation}
    P_d = \dfrac{T_{correct}}{T_{All}}
  \end{equation}
  \item False Alarm Rate($F_a$) : False Alarm Rate is another target-level evaluation metric. It is used to measure the ratio of incorrectly predicted pixels $P_{false}$ to all image pixels $P_{ALL}$. The definition of $F_a$ is as follows:
  \begin{equation}
    F_a = \dfrac{P_{false}}{P_{All}}
  \end{equation}
\end{itemize}

\subsection{Experimental Results}

We compared state-of-the-art methods with our method on SIRST and NUDT-SIRST. And three metrics, IoU, $P_d$, and $F_a$, were used for evaluation. As shown in Tab. ~\ref{tab:exp_res}, our method achieved the best results for the IoU, $P_d$, $F_a$ metrics on SIRST and NUDT-SIRST.
To highlight the superior performance of our method, we visualize our method with other detection models. To highlight the superior performance of our method, we use a red dashed box to circle the target area and zoom in to show it.
To more clearly compare the difference between the model detection results and ground truth, we use red pixels to indicate the difference between the method and ground truth. In addition, we marked the regions of model error detection with yellow dashed boxes.

As shown in Fig.~\ref{fig:visual}, we visualize the performance comparison between the state-of-the-art method and our method on general and difficult samples. Where difficult samples are uncommon instances in the dataset. They usually have large dimensions, multiple objects or irregular contours. General samples are usually small, singular, and have regular shapes.
ACM and ALC-Net had many false alarms. UIU-Net and DNA-Net ~\cite{li2022dense} show fewer wrongly predicted regions, however, they fail to accurately detect the target in some of the difficult samples (the last two samples). 
In contrast, detection model trained with the samples generated by Diff-Mosaic performed well in predictions dealing with difficult samples.
This suggests that Diff-Mosaic can generate more challenging samples for discriminative training in the detection model.

\begin{table}[]
\centering
\caption{We show the number of parameters required and the inference time per sample required for different CNN-based methods.}
\begin{tabular}{ccc}
\hline
Model   & Params(M) & Inference times(s) \\ \hline
ALC-Net & 0.38   & 40.93            \\
ACM-Net & 0.29   & 18.53            \\
DNA-Net & 4.7    & 43.42            \\
UIU-Net & 50.54  & 33.98            \\ \hline
Ours    & 4.7   & 43.42            \\ \hline
\end{tabular}

\label{tab:parm}
\end{table}

As shown in Tab.~\ref{tab:parm}, we list the computational parameter count and inference time of CNN-based methods. Our approach generates augmented samples using pixel-prior and diffusion-prior mechanisms. This process happens during data generation and produces one sample every 10 seconds without affecting training and inference times.  Thus our method is competitive with other methods in terms of the parameter number and inference time.

\subsection{Effects of the Scale in Diff-Mosaic Augmentation }

\begin{table}[]
    \centering
    \caption{
    We conducted experiments using a baseline detection network on the NUDT-SIRST dataset to compare the effect of training with different numbers of synthetic data on accuracy and detection performance. We compared the performance metrics of 125, 250, and 400 augmentation data trained in combination with real data. It can be seen that all three metrics of the model trained with combined training data are better than the model trained with real training data.}
    \begin{tabular}{c|cccc}
    \hline
    real          &  663  &  663  &  663  &  663      \\ \hline 
                         synthetic       & 0   & 125  & 250  & 400      \\ \hline 
                         total         &  663  &  788  & 913  & 1063      \\ \hline
     IoU$ \left( 1\times 10^{-2} )\right \uparrow$ & 88.38 & 90.11 & 90.21 & \textbf{91.18} \\ \hline 
                         $P_d \left( 1\times 10^{-2} )\right \uparrow$  & 97.99 & 98.28 & 98.52 & \textbf{99.47} \\ \hline 
                         $F_a \left( 1\times 10^{-2} )\right\downarrow$ & 4.04 & 3.24 & 3.1  & \textbf{1.91} \\ \hline
    \end{tabular}

\label{tab:num}
\end{table}

We analyzed the effect of different numbers of samples generated by Diff-Mosaic as augmentation samples on the performance of the detection model. As shown in Tab. ~\ref{tab:num}, we compare the performance of the models trained with real data alone and when augmented with 125, 250, and 400 additional samples. With an increase in the number of augmentation samples, the performance of the model, as measured by the $F_a$, $P_d$ and IoU metrics, shows an improvement. This demonstrates the diversity and realism of the augmented samples generated by Diff-Mosaic can improve the robustness of the detection model.

\begin{figure}[ht]
\centering
\includegraphics[width=0.8\linewidth]{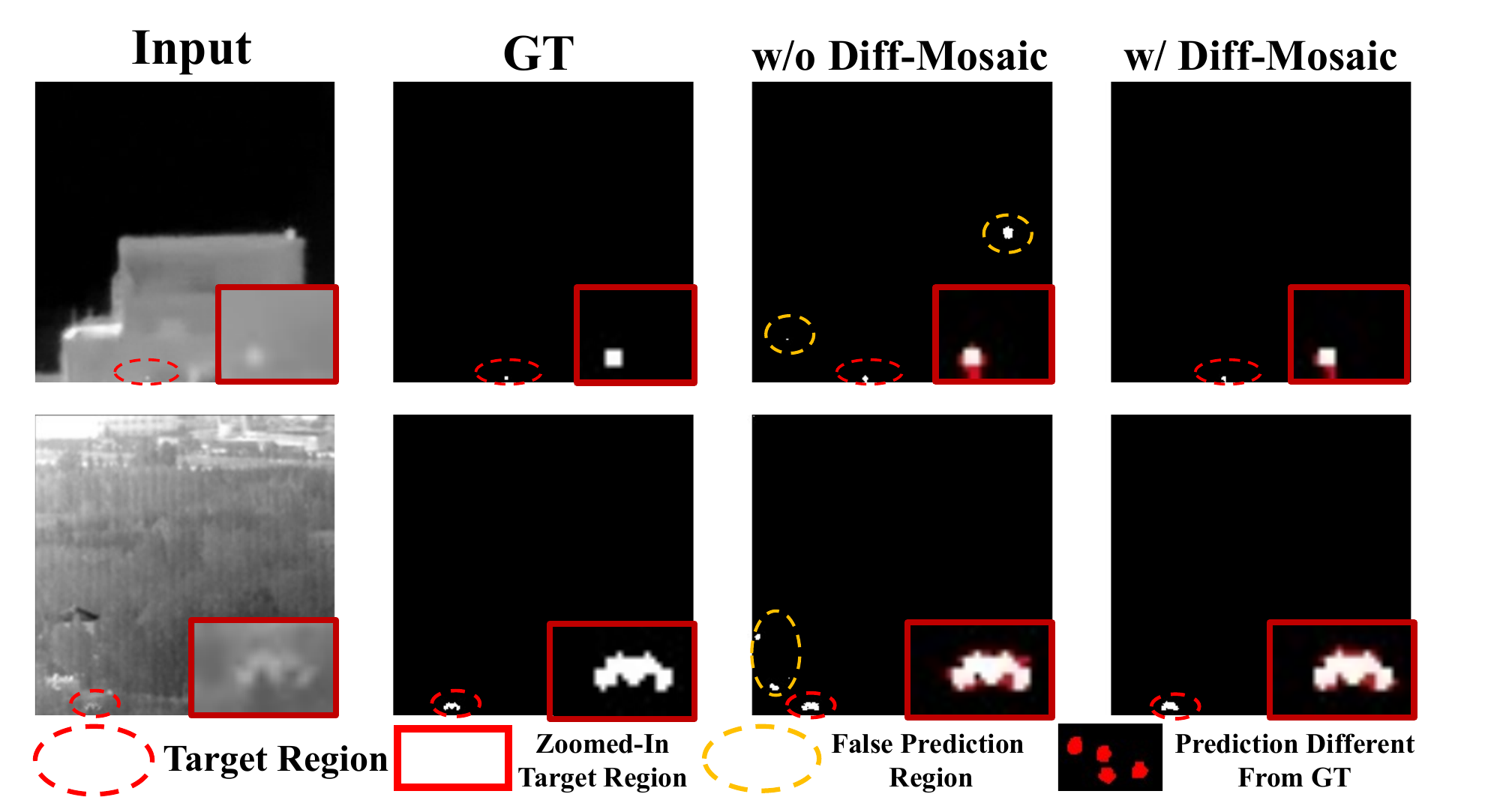}
\caption{Visualization results. We compared the visualization results of `w/o Diff-Mosaic' and `w/ Diff-Mosaic' on UIU-Net, and we can see that UIU-Net has fewer false alarms after training with augmented and broadened samples, and is more accurate in detecting small sample targets.}
\label{fig:UIU}
\end{figure}

\subsection{Ablation Study}

\begin{table*}[ht!]
\centering
\caption{
    Ablation study. We compare the performance gap between each part of our method and the traditional Mosaic. It can be seen that the augmented samples of Mosaic do not provide much improvement in performance. The basic detection network trained using Pixel-Prior generated samples is improved in three evaluation metrics. The basic detection network trained with `w/ Pixel-Prior+Diff-Mosaic' generated samples achieves an impressive promotion.}
    \begin{tabular}{cccc}
    \hline
    \multirow{2}{*}{}          & \multicolumn{3}{c}{NUDT-SIRST(Tr=50\%)}              \\ \cline{2-4} 
                   & IoU( $1\times 10^{-2}$)$\uparrow$ & $P_d$ ( $1\times 10^{-2}$)$\uparrow$ & $F_a$ ( $1\times 10^{-6}$)$\downarrow$ \\ \hline
baseline               & 88.38        & 97.99        & 4.04         \\
w/ Mosaic   & 89.3         & 98.36        & 2.91         \\
w/ Pixel-Prior       & 91.01         & 98.91        & 2.13         \\
w/ Pixel-Prior+ Diff-Prior & \textbf{91.18(\textcolor{red}{+0.17})}         & \textbf{99.47(\textcolor{red}{+0.56})}        &\textbf{ 1.91(\textcolor{blue}{-0.22})}         \\ \hline
\end{tabular}
    
    \label{tab:ab_table}
\end{table*}

The ablation studies in this section were performed on NUDT-SIRST using the baseline detection Network. We compare the performance of models trained with Mosaic, with Pixel-Prior (`w/ Pixel-Prior'), and with Diff-Mosaic (`w/ Pixel-Prior+Diff-Prior'). 
The comparative results are shown in Tab.~\ref{tab:ab_table}. Due to the lack of realism and diversity of the samples generated by `w/Mosaic', the improvement of the detection network performance is very limited. And `w/ Pixel-Prior' harmonizes the augmented samples, making them more realistic and therefore improving on all three performance metrics. At last, 'w/ Pixel-Prior+Diff-Prior' introduces real-world information and generates diverse and realistic augmented samples. Aided by 'w/ Pixel-Prior+Diff-Prior', the detection network shows significant improvement in all three performance metrics.

\subsection{Effects of Diff-Mosaic}

\begin{table}[]
\centering
\caption{Quantitative results of UIU-Net with and without diff-mosaic. It can be seen that the results of 'w/ Diff-Mosaic' are significantly improved compared to 'w/o Diff-Mosaic'.}
\begin{tabular}{ccc}
\hline                        
          & IoU$ \left( 1\times 10^{-2} )\right \uparrow$  & $F_a \left( 1\times 10^{-6} )\right  \downarrow$ \\ \hline
w/o Diff-Mosaic      & 89       
            & 6.02           \\
w/ Diff-Mosaic  & \textbf{91.91(\textcolor{red}{+2.91})}          &\textbf{ 1.40(\textcolor{blue}{-4.62})}       \\ \hline
\end{tabular}

\label{tab:UIU}
\end{table}

\begin{table}[]
\centering
\caption{Quantitative results of DNA-Net with and without diff-mosaic. It can be seen that the results of 'w/ Diff-Mosaic' are significantly improved compared to 'w/o Diff-Mosaic'.}
\begin{tabular}{ccc}
\hline 
         & IoU$ \left( 1\times 10^{-2} )\right \uparrow$  & $F_a \left( 1\times 10^{-6} )\right  \downarrow$ \\ \hline
w/o Diff-Mosaic      & 88.38                         & 4.04           \\
w/ Diff-Mosaic  & \textbf{91.18(\textcolor{red}{+2.8}) }              & \textbf{1.91(\textcolor{blue}{-2.13})}        \\ \hline
\end{tabular}
\label{tab:DNA}
\end{table}

To demonstrate the effectiveness of the augmentation samples generated by Diff-Mosaic, we used the generated results of Diff-Mosaic as augmentation samples for training on different detection models. In this section, we compare the effects of training without (`w/o Diff-Mosaic') and with Diff-Mosaic's (`w/ Diff-Mosaic') augmented samples using UIU-Net and DNA-Net as detection models.
As shown in Tab.~\ref{tab:UIU}, we compare the performance of the UIU-Net model with and without the augmented samples. To highlight the gaps, we use red bold font to indicate how much `w/Diff-Mosaic' improves on IoU and blue bold font to indicate how much `w/ Diff-Mosaic' decreases on $F_a$.
It can be seen that the performance of UIUNet improves significantly on the NUDT-SIRST dataset after training with Diff-Mosaic augmented samples. As shown in Fig.~\ref{fig:UIU}, it can be seen that `w/o Diff-Mosaic' has a lot of misdirected targets in its detection results and is inaccurate in detecting the contours of small targets. Furthermore, the detection result of `w/ Diff-Mosaic' has no false detection targets and achieves accurate detection of the contours of small irregular targets.
\begin{table}[]
\centering
\caption{ 
Performance comparison of different image generation methods. 
We have used FID and KID to measure the realism of the generated images. It can be seen that our method achieves the best results.
}
\begin{tabular}{ccc}
\hline
           & FID$\downarrow$   & KID$\downarrow$  \\ \hline
SwinIR~\cite{liang2021swinir}     & 181.39 & 0.167 \\
FeMaSR~\cite{chen2022real}     & 200.47 & 0.191 \\
DiffBir~\cite{lin2023diffbir}    & 188.22 & 0.163 \\ \hline
diffmosaic & 126.01 & 0.076 \\ \hline
\end{tabular}
\label{tab:ir_res}
\end{table}

\begin{figure}[h]
\centering
\includegraphics[width=0.8\linewidth]{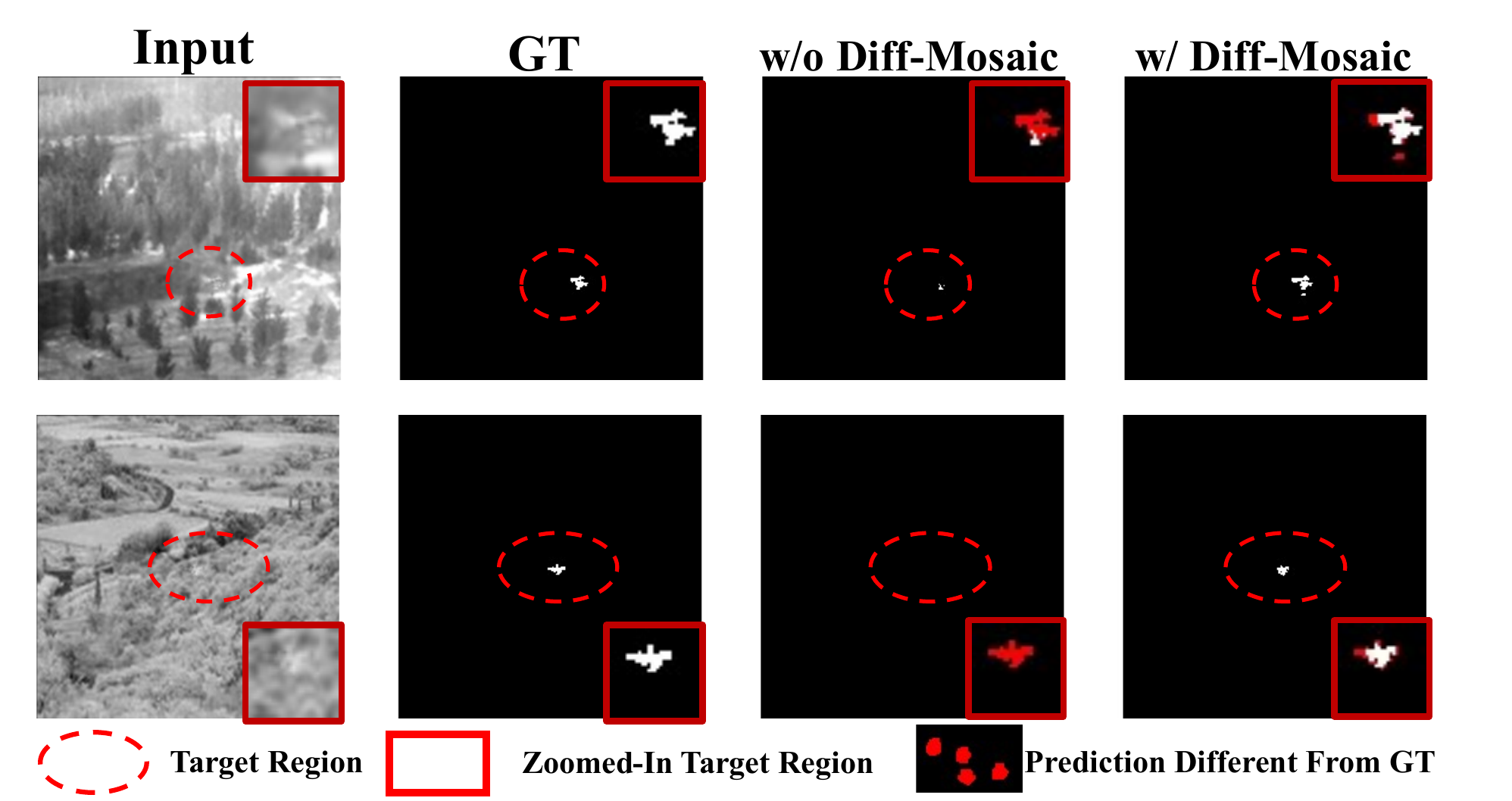}
\caption{Visualization results. We compared the visualization results of `w/o Diff-Mosaic' and `w/ Diff-Mosaic' on DNA-Net.
}
\label{fig:DNA}
\end{figure}

\begin{table*}[t]
\centering
\normalsize
\caption{Applying diffusion prior to  CutMix~\cite{yun2019cutmix}. It can be seen that the augmented samples generated by `w/CutMix + our diffusion prior' can stimulate the potential of the detection model more than `w/CutMix'.}
\begin{tabular}{cccc}
\hline
    Method                    & IoU($1 \times 10^{-2}$) $\uparrow$  & $P_d$($1 \times 10^{-2}$) $\uparrow$    & $F_a (1 \times 10^{-6})$ $\downarrow$   \\ \hline
    Baseline model                  & 88.38 & 97.99  & 4.04  \\
    w/ CutMix                 & 89.24 & 98.62 & 6.25 \\ \hline
    w/CutMix + our diffusion prior & 90.11 & 99.,15 & 2.04  \\ \hline
    \end{tabular}
    
\label{tab:cutmix}
\end{table*}

\begin{figure}[t]
    \centering
    \includegraphics[width=0.85\linewidth]{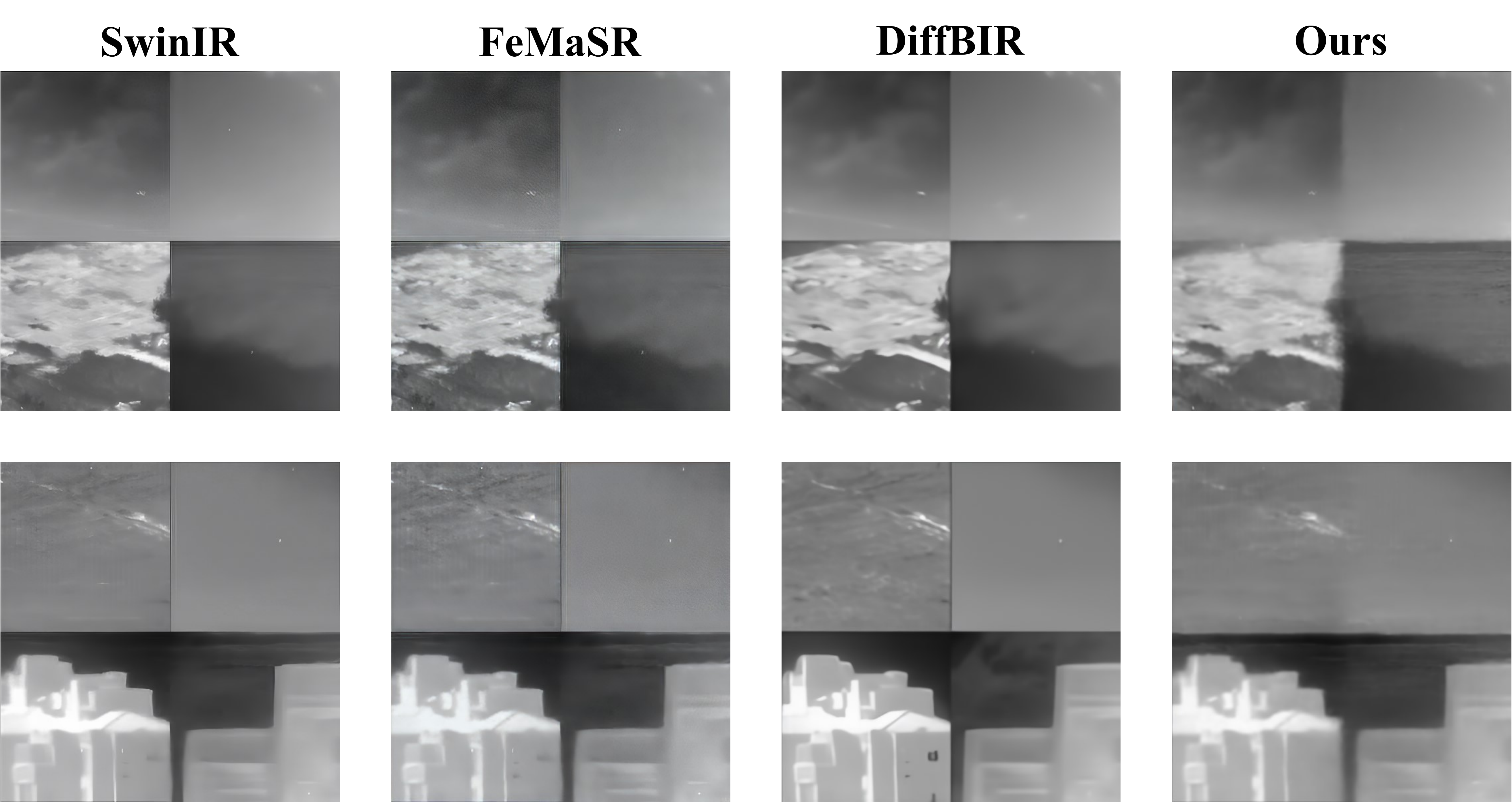}
    \caption[something short]{Visualization results of different image generation methods. We visualized the generation results of different generation methods. It can be seen that our method integrates the images well and makes them more coherent than other methods.}
    \label{fig:ir}
\vspace{-3mm}
\end{figure}

As shown in Tab.~\ref{tab:DNA}, we compared the performance of the DNA-Net model with and without the augmented samples generated by Diff-Mosaic. The results indicate a significant improvement in detection performance on the NUDT-SIRST dataset when using augmented samples.
As depicted in Fig.~\ref{fig:DNA}, it is evident that the `w/o Diff-Mosaic" method fails to detect the target in challenging test samples, while the `w/ Diff-Mosaic' method accurately identifies the target with high precision.
To show the realism of the images we generate, we input the results of the pixel-prior machine into state-of-the-art image generation techniques, including SwinIR~\cite{liang2021swinir}, FemaSR~\cite{chen2022real} and DiffBIR~\cite{lin2023diffbir}, and compare their generated results with our method. 
As shown in Fig.~\ref{fig:ir}, it can be observed that augmented samples generated by SwinIR, FeMaSR, and DiffBIR fail to integrate the Mosaic strategy well. On the contrary, the augmented samples generated by our diff-mosaic demonstrate a seamless coherence in image distribution and a high level of realism.
Additionally, to measure the realism of the generated images, we utilized KID~\cite{binkowski2018demystifying} and FID~\cite{heusel2017gans} metrics to measure the disparity between them and real infrared images. As shown in tab.~\ref{tab:ir_res}, our method outperforms others in terms of FID and KID, indicating a significant advantage in enhancing realism.

\subsection{Effects of Diffusion-Prior}
\begin{figure}[h]
    \centering
    \includegraphics[width=0.85\linewidth]{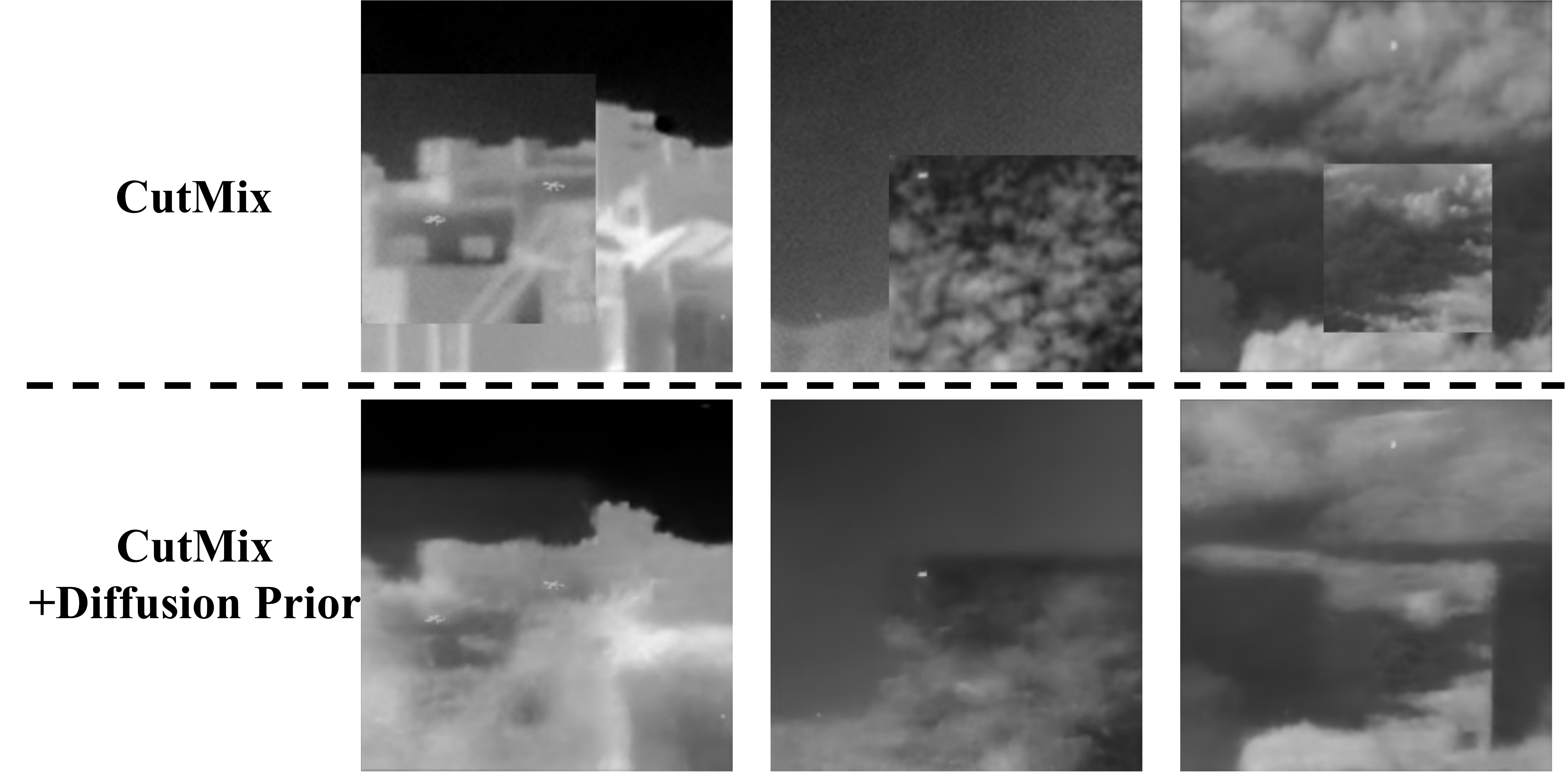}
    \caption[something short]{We visualized the augmented samples generated by the CutMix~\cite{yun2019cutmix} method with the augmented samples generated in combination with our method. 

    }
    \label{fig:cut}
\vspace{-3mm}
\end{figure}

To demonstrate the effectiveness of diffusion-prior machine, we combined it with CutMix~\cite{yun2019cutmix}. Subsequently, we generated 400 augmented samples on NUDT-SIRST and input them into the detection network for training.
As depicted in Fig.~\ref{fig:cut}, we demonstrate the augmented samples from the original CutMix (`w/ CutMix') and the combination with our method (`w/  CutMix + diffusion prior'). It is evident that the results generated by the original CutMix lack coherence and realism. In contrast, the results generated by `w/ CutMix + diffusion prior' exhibit greater coherence. 
To validate the effectiveness of our diffusion prior, we compared the performance of the augmented samples generated by `w/ CutMix' and `w/ CutMix + our diffusion prior'. As shown in Tab.~\ref{tab:cutmix}, it can be observed that 'w/ CutMix' shows a negligible improvement in the network. In contrast, 'w/ CutMix + diffusion prior' introduces real-world information, leading to a noticeable improvement in the network's detection capability.

\section{Conclusion}
\label{sec:label}
In this paper, a novel diffusion-based data augmentation method is proposed to solve the problems of lack of realism and lack of diversity of existing data augmentation methods.
Our method generates harmonized augmented samples based on the image pixel distribution, making the samples more realistic. We then use a powerful diffusion model prior that introduces real-world information by resampling the sample. 
This resampling makes the generated samples realistic and diverse. Extensive experiments show that our method effectively improves the performance of the model. In addition, ablation experiments show that subcomponents of our method are effective and that the augmented samples generated by our method can improve the performance of different baseline detection networks.

\bibliographystyle{IEEEtran}
\bibliography{egbib}

\begin{thebibliography}{10}
\providecommand{\url}[1]{#1}
\csname url@samestyle\endcsname
\providecommand{\newblock}{\relax}
\providecommand{\bibinfo}[2]{#2}
\providecommand{\BIBentrySTDinterwordspacing}{\spaceskip=0pt\relax}
\providecommand{\BIBentryALTinterwordstretchfactor}{4}
\providecommand{\BIBentryALTinterwordspacing}{\spaceskip=\fontdimen2\font plus
\BIBentryALTinterwordstretchfactor\fontdimen3\font minus \fontdimen4\font\relax}
\providecommand{\BIBforeignlanguage}[2]{{%
\expandafter\ifx\csname l@#1\endcsname\relax
\typeout{** WARNING: IEEEtran.bst: No hyphenation pattern has been}%
\typeout{** loaded for the language `#1'. Using the pattern for}%
\typeout{** the default language instead.}%
\else
\language=\csname l@#1\endcsname
\fi
#2}}
\providecommand{\BIBdecl}{\relax}
\BIBdecl

\bibitem{bochkovskiy2020yolov4}
A.~Bochkovskiy, C.-Y. Wang, and H.-Y.~M. Liao, ``Yolov4: Optimal speed and accuracy of object detection,'' \emph{arXiv preprint arXiv:2004.10934}, 2020.

\bibitem{xiao2023ediffsr}
Y.~Xiao, Q.~Yuan, K.~Jiang, J.~He, X.~Jin, and L.~Zhang, ``Ediffsr: An efficient diffusion probabilistic model for remote sensing image super-resolution,'' \emph{IEEE Transactions on Geoscience and Remote Sensing}, 2023.

\bibitem{hu2023cycmunet}
M.~Hu, K.~Jiang, Z.~Wang, X.~Bai, and R.~Hu, ``Cycmunet+: Cycle-projected mutual learning for spatial-temporal video super-resolution,'' \emph{IEEE Transactions on Pattern Analysis and Machine Intelligence}, 2023.

\bibitem{xiao2304local}
Y.~Xiao, Q.~Yuan, K.~Jiang, X.~Jin, J.~He, L.~Zhang, and C.~Lin, ``Local-global temporal difference learning for satellite video super-resolution. arxiv 2023,'' \emph{arXiv preprint arXiv:2304.04421}.

\bibitem{deng2016small}
H.~Deng, X.~Sun, M.~Liu, C.~Ye, and X.~Zhou, ``Small infrared target detection based on weighted local difference measure,'' \emph{IEEE Transactions on Geoscience and Remote Sensing}, vol.~54, no.~7, pp. 4204--4214, 2016.

\bibitem{jiang2024mutual}
K.~Jiang, Q.~Wang, Z.~An, Z.~Wang, C.~Zhang, and C.-W. Lin, ``Mutual retinex: Combining transformer and cnn for image enhancement,'' \emph{IEEE Transactions on Emerging Topics in Computational Intelligence}, 2024.

\bibitem{chen2023towards}
X.~Chen, J.~Pan, J.~Dong, and J.~Tang, ``Towards unified deep image deraining: A survey and a new benchmark,'' \emph{arXiv preprint arXiv:2310.03535}, 2023.

\bibitem{jiangfmrnet}
K.~Jiang, J.~Jiang, X.~Liu, X.~Xu, and X.~Ma, ``Fmrnet: Image deraining via frequency mutual revision.''

\bibitem{rawat2020review}
S.~S. Rawat, S.~K. Verma, and Y.~Kumar, ``Review on recent development in infrared small target detection algorithms,'' \emph{Procedia Computer Science}, vol. 167, pp. 2496--2505, 2020.

\bibitem{rivest1996detection}
J.-F. Rivest and R.~Fortin, ``Detection of dim targets in digital infrared imagery by morphological image processing,'' \emph{Optical Engineering}, vol.~35, no.~7, pp. 1886--1893, 1996.

\bibitem{bai2010analysis}
X.~Bai and F.~Zhou, ``Analysis of new top-hat transformation and the application for infrared dim small target detection,'' \emph{Pattern Recognition}, vol.~43, no.~6, pp. 2145--2156, 2010.

\bibitem{deshpande1999max}
S.~D. Deshpande, M.~H. Er, R.~Venkateswarlu, and P.~Chan, ``Max-mean and max-median filters for detection of small targets,'' in \emph{Signal and Data Processing of Small Targets 1999}, vol. 3809.\hskip 1em plus 0.5em minus 0.4em\relax SPIE, 1999, pp. 74--83.

\bibitem{chen2013local}
C.~P. Chen, H.~Li, Y.~Wei, T.~Xia, and Y.~Y. Tang, ``A local contrast method for small infrared target detection,'' \emph{IEEE transactions on geoscience and remote sensing}, vol.~52, no.~1, pp. 574--581, 2013.

\bibitem{han2020infrared}
J.~Han, S.~Moradi, I.~Faramarzi, H.~Zhang, Q.~Zhao, X.~Zhang, and N.~Li, ``Infrared small target detection based on the weighted strengthened local contrast measure,'' \emph{IEEE Geoscience and Remote Sensing Letters}, vol.~18, no.~9, pp. 1670--1674, 2020.

\bibitem{han2019local}
J.~Han, S.~Moradi, I.~Faramarzi, C.~Liu, H.~Zhang, and Q.~Zhao, ``A local contrast method for infrared small-target detection utilizing a tri-layer window,'' \emph{IEEE Geoscience and Remote Sensing Letters}, vol.~17, no.~10, pp. 1822--1826, 2019.

\bibitem{gao2013infrared}
C.~Gao, D.~Meng, Y.~Yang, Y.~Wang, X.~Zhou, and A.~G. Hauptmann, ``Infrared patch-image model for small target detection in a single image,'' \emph{IEEE transactions on image processing}, vol.~22, no.~12, pp. 4996--5009, 2013.

\bibitem{dai2021asymmetric}
Y.~Dai, Y.~Wu, F.~Zhou, and K.~Barnard, ``Asymmetric contextual modulation for infrared small target detection,'' in \emph{Proceedings of the IEEE/CVF Winter Conference on Applications of Computer Vision}, 2021, pp. 950--959.

\bibitem{wang2019miss}
H.~Wang, L.~Zhou, and L.~Wang, ``Miss detection vs. false alarm: Adversarial learning for small object segmentation in infrared images,'' in \emph{Proceedings of the IEEE/CVF International Conference on Computer Vision}, 2019, pp. 8509--8518.

\bibitem{dai2021attentional}
Y.~Dai, Y.~Wu, F.~Zhou, and K.~Barnard, ``Attentional local contrast networks for infrared small target detection,'' \emph{IEEE Transactions on Geoscience and Remote Sensing}, vol.~59, no.~11, pp. 9813--9824, 2021.

\bibitem{li2022dense}
B.~Li, C.~Xiao, L.~Wang, Y.~Wang, Z.~Lin, M.~Li, W.~An, and Y.~Guo, ``Dense nested attention network for infrared small target detection,'' \emph{IEEE Transactions on Image Processing}, vol.~32, pp. 1745--1758, 2022.

\bibitem{wu2022uiu}
X.~Wu, D.~Hong, and J.~Chanussot, ``Uiu-net: U-net in u-net for infrared small object detection,'' \emph{IEEE Transactions on Image Processing}, vol.~32, pp. 364--376, 2022.

\bibitem{yun2019cutmix}
S.~Yun, D.~Han, S.~J. Oh, S.~Chun, J.~Choe, and Y.~Yoo, ``Cutmix: Regularization strategy to train strong classifiers with localizable features,'' in \emph{Proceedings of the IEEE/CVF international conference on computer vision}, 2019, pp. 6023--6032.

\bibitem{zhang2017mixup}
H.~Zhang, M.~Cisse, Y.~N. Dauphin, and D.~Lopez-Paz, ``mixup: Beyond empirical risk minimization,'' \emph{arXiv preprint arXiv:1710.09412}, 2017.

\bibitem{fei2023generative}
B.~Fei, Z.~Lyu, L.~Pan, J.~Zhang, W.~Yang, T.~Luo, B.~Zhang, and B.~Dai, ``Generative diffusion prior for unified image restoration and enhancement,'' in \emph{Proceedings of the IEEE/CVF Conference on Computer Vision and Pattern Recognition}, 2023, pp. 9935--9946.

\bibitem{chen2023controlstyle}
J.~Chen, Y.~Pan, T.~Yao, and T.~Mei, ``Controlstyle: Text-driven stylized image generation using diffusion priors,'' in \emph{Proceedings of the 31st ACM International Conference on Multimedia}, 2023, pp. 7540--7548.

\bibitem{yu2024hypersinet}
Q.~Yu, W.~Wei, D.~Li, Z.~Pan, C.~Li, and D.~Hong, ``Hypersinet: A synergetic interaction network combined with convolution and transformer for hyperspectral image classification,'' \emph{IEEE Transactions on Geoscience and Remote Sensing}, 2024.

\bibitem{zhang2024deep}
J.~Zhang, Q.~Cui, and X.~Ma, ``Deep evidential remote sensing landslide image classification with a new divergence, multi-scale saliency and an improved three-branched fusion,'' \emph{IEEE Journal of Selected Topics in Applied Earth Observations and Remote Sensing}, 2024.

\bibitem{wu2023three}
G.~Wu, X.~Ning, L.~Hou, F.~He, H.~Zhang, and A.~Shankar, ``Three-dimensional softmax mechanism guided bidirectional gru networks for hyperspectral remote sensing image classification,'' \emph{Signal Processing}, vol. 212, p. 109151, 2023.

\bibitem{jha2023mdfs}
A.~Jha and B.~Banerjee, ``Mdfs-net: Multi-domain few shot classification for hyperspectral images with support set reconstruction,'' \emph{IEEE Transactions on Geoscience and Remote Sensing}, 2023.

\bibitem{miao2023ecae}
W.~Miao, Z.~Xu, J.~Geng, and W.~Jiang, ``Ecae: Edge-aware class activation enhancement for semisupervised remote sensing image semantic segmentation,'' \emph{IEEE Transactions on Geoscience and Remote Sensing}, 2023.

\bibitem{kou2023lw}
R.~Kou, C.~Wang, Y.~Yu, Z.~Peng, M.~Yang, F.~Huang, and Q.~Fu, ``Lw-irstnet: Lightweight infrared small target segmentation network and application deployment,'' \emph{IEEE Transactions on Geoscience and Remote Sensing}, 2023.

\bibitem{gao2023integrating}
K.~Gao, A.~Yu, X.~You, W.~Guo, K.~Li, and N.~Huang, ``Integrating multiple sources knowledge for class asymmetry domain adaptation segmentation of remote sensing images,'' \emph{IEEE Transactions on Geoscience and Remote Sensing}, 2023.

\bibitem{shamsolmoali2023efficient}
P.~Shamsolmoali, J.~Chanussot, H.~Zhou, and Y.~Lu, ``Efficient object detection in optical remote sensing imagery via attention-based feature distillation,'' \emph{IEEE Transactions on Geoscience and Remote Sensing}, 2023.

\bibitem{ronneberger2015u}
O.~Ronneberger, P.~Fischer, and T.~Brox, ``U-net: Convolutional networks for biomedical image segmentation,'' in \emph{Medical image computing and computer-assisted intervention--MICCAI 2015: 18th international conference, Munich, Germany, October 5-9, 2015, proceedings, part III 18}.\hskip 1em plus 0.5em minus 0.4em\relax Springer, 2015, pp. 234--241.

\bibitem{lin2017feature}
T.-Y. Lin, P.~Doll{\'a}r, R.~Girshick, K.~He, B.~Hariharan, and S.~Belongie, ``Feature pyramid networks for object detection,'' in \emph{Proceedings of the IEEE conference on computer vision and pattern recognition}, 2017, pp. 2117--2125.

\bibitem{ho2020denoising}
J.~Ho, A.~Jain, and P.~Abbeel, ``Denoising diffusion probabilistic models,'' \emph{Advances in neural information processing systems}, vol.~33, pp. 6840--6851, 2020.

\bibitem{rombach2022high}
R.~Rombach, A.~Blattmann, D.~Lorenz, P.~Esser, and B.~Ommer, ``High-resolution image synthesis with latent diffusion models,'' in \emph{Proceedings of the IEEE/CVF conference on computer vision and pattern recognition}, 2022, pp. 10\,684--10\,695.

\bibitem{kingma2013auto}
D.~P. Kingma and M.~Welling, ``Auto-encoding variational bayes,'' \emph{arXiv preprint arXiv:1312.6114}, 2013.

\bibitem{lin2023diffbir}
X.~Lin, J.~He, Z.~Chen, Z.~Lyu, B.~Fei, B.~Dai, W.~Ouyang, Y.~Qiao, and C.~Dong, ``Diffbir: Towards blind image restoration with generative diffusion prior,'' \emph{arXiv preprint arXiv:2308.15070}, 2023.

\bibitem{liu2021swin}
Z.~Liu, Y.~Lin, Y.~Cao, H.~Hu, Y.~Wei, Z.~Zhang, S.~Lin, and B.~Guo, ``Swin transformer: Hierarchical vision transformer using shifted windows,'' in \emph{Proceedings of the IEEE/CVF international conference on computer vision}, 2021, pp. 10\,012--10\,022.

\bibitem{liang2021swinir}
J.~Liang, J.~Cao, G.~Sun, K.~Zhang, L.~Van~Gool, and R.~Timofte, ``Swinir: Image restoration using swin transformer,'' in \emph{Proceedings of the IEEE/CVF international conference on computer vision}, 2021, pp. 1833--1844.

\bibitem{chen2022real}
C.~Chen, X.~Shi, Y.~Qin, X.~Li, X.~Han, T.~Yang, and S.~Guo, ``Real-world blind super-resolution via feature matching with implicit high-resolution priors,'' in \emph{Proceedings of the 30th ACM International Conference on Multimedia}, 2022, pp. 1329--1338.

\bibitem{binkowski2018demystifying}
M.~Bi{\'n}kowski, D.~J. Sutherland, M.~Arbel, and A.~Gretton, ``Demystifying mmd gans,'' in \emph{International Conference on Learning Representations}, 2018.

\bibitem{heusel2017gans}
M.~Heusel, H.~Ramsauer, T.~Unterthiner, B.~Nessler, and S.~Hochreiter, ``Gans trained by a two time-scale update rule converge to a local nash equilibrium,'' \emph{Advances in neural information processing systems}, vol.~30, 2017.

\end{thebibliography}
\end{document}